\documentclass[letterpaper, 10 pt, conference]{ieeeconf}  

\IEEEoverridecommandlockouts                              

\usepackage{hyperref}       
\usepackage{url}            
\usepackage{booktabs}       
\usepackage{amsfonts}       
\usepackage{amsmath}
\usepackage{color}
\usepackage{algorithm}
\usepackage{algorithmic}
\usepackage{float}

\usepackage{caption}
\usepackage{subcaption}
\usepackage{graphicx}

\title{\LARGE \bf
Autonomous Highway Driving using Deep Reinforcement Learning
}

\author{Subramanya Nageshrao$^{1}$ and Eric Tseng$^{2}$ and Dimitar Filev$^{2}$
\thanks{$^{1}$ Ford Greenfield Labs, 3251 Hillview Ave, Palo Alto, CA 94304, USA 
        {\tt\small snageshr@ford.com}.}%
\thanks{$^{2}$ Ford Research and Innovation Center, 2101 Village Road, Dearborn, MI 48124, USA,
        {\tt\small htseng@ford.com},
        {\tt\small dfilev@ford.com}.}%
}

\begin{document}

\maketitle
\thispagestyle{empty}
\pagestyle{empty}

\begin{abstract}
The operational space of an autonomous vehicle (AV) can be diverse and vary significantly. This may lead to a scenario that was not postulated in the design phase. Due to this, formulating a rule based decision maker for selecting maneuvers may not be ideal. Similarly, it may not be effective to design an a-priori cost function and then solve the optimal control problem in real-time. In order to address these issues and to avoid peculiar behaviors when encountering unforeseen scenario, we propose a reinforcement learning (RL) based method, where the ego car, i.e., an autonomous vehicle, learns to make decisions by directly interacting with simulated traffic. The decision maker for AV is implemented as a deep neural network providing an action choice for a given system state. 

In a critical application such as driving, an RL agent without explicit notion of safety may not converge or it may need extremely large number of samples before finding a reliable policy. To best address the issue, this paper incorporates reinforcement learning with an additional short horizon safety check (SC). In a critical scenario, the safety check will also provide an alternate safe action to the agent provided if it exists. This leads to two novel contributions. First, it generalizes the states that could lead to undesirable ``near-misses" or ``collisions ". Second, inclusion of safety check can provide a safe and stable training environment. This significantly enhances learning efficiency without inhibiting meaningful exploration to ensure safe and optimal learned behavior.

We demonstrate the performance of the developed algorithm in highway driving scenario where the trained AV encounters varying traffic density in a highway setting.
 
\end{abstract}

\section{Introduction}
Reinforcement learning (RL), is a branch of artificial intelligence where an agent learns an optimal control strategy by interacting with the environment. Over the last few years there are many monumental success stories in RL. The most spectacular and well-known results involve state-of-the-art performance in many challenging tasks such as video games \cite{mnih2015human}, decision making in an extremely complex environments  such as game of Go \cite{silver2017mastering}, and  continuous control of physical systems \cite{lillicrap2015continuous,mnih2016asynchronous}.
 
In recent years autonomous driving has been a subject of great interest among both researchers and general public. Autonomous vehicles (AVs) have enormous potential not only economically but also for enhancing mobility options, and potentially reducing carbon footprint \cite{paden2016survey}. As in any robotic system, autonomous driving also involves decision making. A typical implementation of AV decision making is done by solving a number of path planning problems. They are formulated as a set of optimization problems,  subject to various system and environmental constraints. The solution corresponding to the lowest cost is often chosen as reference to the low-level controller \cite{schwarting2018planning}. This approach is computationally intensive, additionally due to the real time constraints on the solver,  the obtained reference path may not result in a comfortable vehicle motion. Compare this to human driving, where a driver perceives the surrounding traffic and makes a maneuver decision, such as change lane, maintain speed, or brake etc. Based on this decision, the driver then proceeds with steering and throttle/brake pedal actuation to perform the maneuver with a smooth path. This approach fits well within the realm of the RL framework.        

In this work we develop a Deep Reinforcement Learning (DRL) agent for highway driving.  The DRL agent is designed by training an ego vehicle (EV) to learn a driving policy by directly interacting with diverse simulated traffic. The foundation of our DRL agent is a modified version of the  double deep $Q$-network (DDQN) algorithm that was
first discussed in \cite{van2016deep,hessel2018rainbow}. DDQN is considered as one of the state-of-the art RL algorithm for discrete state and continuous action space optimal problems. As detailed in \cite{van2016deep}, DDQN not only rectifies overestimation problem of DQN \cite{mnih2015human} but also improves performance across many benchmark problems. In this work we customize DDQN from \cite{van2016deep} to include the following distinctive features.  
\begin{itemize}
\item First, we augment the DDQN learning agent with a short-horizon safety check. In a critical situation, if the original DDQN action choice was deemed unsafe then the safety check will replace it with an alternate safe action, provided there is a safe alternative. For training, Q-learning often relies on an exploration policy, such as $\epsilon$-greedy, as we have experienced and as one may anticipate, arbitrary exploration without constraints could frequently lead to scenarios involving collisions, and in turn, simulation resets. The safety check augmentation significantly improves the learning efficiency in avoiding frequent resets during  training. Additionally during inference phase the safety check can be used as an override authority whenever a non-safe maneuver was chosen by the algorithm. This rare situation may occur for example, due to the use of function approximation such as neural networks. The safety rules is based on kinematic constraints so as to minimize the collision risks, they are elaborated in section~\ref{sec::DQNArchitecture} and are similar to the ones proposed in \cite{shalev2017formal}.

\item  Second, we keep the situations that invoke the safety violation in a “collision buffer”. As such, the safety check facilitates a constrained exploration during learning which enhances learning efficiency and achieves stable learning \cite{garcia2015comprehensive,alshiekh2017safe}. Use of two buffers is a simpler version of prioritized experience replay as discussed in \cite{schaul2015prioritized}.

\item Third, the inclusion of \emph{safety controller} during inference phase leads to an interesting scenario, where by using new information the learned agent can be re-trained thus resulting in continuous adaptation. The new information can be potentially obtained from real-world data by deploying the initially trained agent in multiple vehicles. 
\end{itemize}
These changes were found to be essential for both stability and efficiency of the DRL agent and will be discussed in detail in Section.~\ref{sec::evalLearnedControl} of the paper.   

We demonstrate the usability of the developed algorithm for highway driving of an AV with varying traffic density in a simulation environment.  In summary, the key contributions of this work are:
\begin{enumerate}
\item We develop a DRL agent and a corresponding control architecture for highway autonomous driving using deep $Q$-network as a decision maker in conjunction with standard state feedback controllers.
\item We integrate the semantic information of the rules of the road in the form of a \emph{short-horizon safety check} complementing with the statistical inference mechanism used in the  DDQN algorithm. 
\item We demonstrated the capability of the developed DRL agent in an adaptive framework in presence of new data. 
\end{enumerate}

The rest of paper is organized as follows. In Section~\ref{sec::rlprelim}, a preliminary introduction on RL for decision making and control is presented. Following that, in Section.~\ref{sec::DQNArchitecture}, the proposed DDQN based algorithm along with the DRL based control architecture is explained. The implementation of the DRL agent for highway driving of an autonomous vehicle in an varying traffic density is given in Section.~\ref{sec::evalLearnedControl}. Finally, Section.~\ref{sec:conclusion} concludes the paper with a note on our future research.

\section{Preliminaries: Deep learning for control} \label{sec::rlprelim}

In this section we provide a brief theoretical background on decision making and control using deep reinforcement learning. Additionally, we elaborate on the main motivation for the control architecture presented in this article.

\subsection{Reinforcement learning}
In reinforcement learning (RL) an agent learns an optimal behavior with respect to a cost function by directly interacting with the system. Generally, RL problem is formulated as a Markov decision process (MDP) which is formally defined by a tuple $\langle \mathcal{S}, \mathcal{A}, \mathcal{T}, \mathcal{R} \rangle$. Where $\mathcal{S}\in \mathbb{R}^n$ is the state-space, $\mathcal{A}$ is the action space, $\mathcal{T}:\mathcal{S}\times \mathcal{A}\rightarrow \mathcal{S}$ is the state transformation function, and $\mathcal{R}:\mathcal{S}\times \mathcal{A} \times \mathcal{S} \rightarrow \mathbb{R}$ is the user defined reward function. 

At each time step $t$ the learning agent selects an action $a_t$ according to some policy $\pi$ based on the current system state $s_t$ , i.e., $a_t = \pi(s_t)$. For discrete action space $a_t$ will be from a set of feasible actions in  $\mathcal{A}$. On applying this action the system's state $s_t$ transitions to a new state $s_{t+1}$,  along with this the environment provides a scalar reward $r_{t+1}$. This process is repeated either until $N_\mathrm{s}$ samples or the agent reaches a terminal state, this is called a learning episode. After the termination, the learning episode is restarted, this step is repeated either until $N_\mathrm{e}$ episodes or convergence. In RL, it is assumed that the system under consideration satisfies the Markovian property \cite{sutton1998reinforcement}.

The goal in RL is to learn a policy $\pi$ so as to maximize the total accumulated reward termed as return $R_\mathrm{t} = \sum_{k=0}^{\infty} \gamma^k r_{t+k}$ where the scalar constant $\gamma \in \left( 0,1 \right]$ is called the discount factor. For discrete action space the optimal policy is obtained by solving for the optimal $Q$ function, called action value function. An optimal $Q$ function satisfies the Bellman equation:
\begin{equation}\label{Sec2Eq1:BellmanQ}
Q^{\pi_\mathrm{opt}}(s,a) =  Q^{*}(s,a) = \mathbb{E}_{s^{'}} \left[ r+\gamma \max_{a^{'}} Q^{*}(s^{'},a^{'}) | (s,a) \right]. 
\end{equation} 

Any action $a$ that satisfies \eqref{Sec2Eq1:BellmanQ} in state $s$ will be an optimal action in that state. Since last few years RL has had a significant success \cite{mnih2015human,silver2017mastering,hessel2017rainbow,mnih2016asynchronous,lillicrap2015continuous}. This can be mainly attributed to augmenting RL with deep learning techniques \cite{arulkumaran2017deep}.

\subsection{Decision making and low-level actuation}
In this section, inspired by an example from \cite{sutton1998reinforcement} (see Chapter 1.2), we elaborate on the
 difference between decision making ($\mathcal{D}$'s) and low level actuation ($\mathcal{C}$'s) : Phil wants to have breakfast, he may chose between having cereal or a bagel ($\mathcal{D}_1$), after deciding he will either walk to the cupboard or to the counter ($\mathcal{C}_1$). Based on $\mathcal{D}_1$ he will either pick a choice of cereal ($\mathcal{D}_2$) then walk to fridge ($\mathcal{C}_2$) to get milk or chose a bagel ($\mathcal{D}_3$) and walk to the toaster ($\mathcal{C}_3$), etc. This entire process involves a series of decision making followed by low-level actuation. A high-level decisions (i.e., $\mathcal{D}$'s) are decided by reward function such as pleasure or getting enough nutrition etc. Whereas low level actuation is characterized by a series of hand-eye coordination.

Any RL methodology that tries to solve both decision making and actuation simultaneously may require a large amount of training data. In this work we simplify this process by having a clear hierarchy between high-level decision making and low level actuation. Whereas RL is used to solve the decision making problem, classical feedback control method such as PID is used for low-level actuation.  In a way the methodology presented here has considerable overlap with hierarchical RL where both decision making and control are learned simultaneously subject to a set of related but different cost functions \cite{frans2017meta}. For training the decision making agent we use a modified DDQN algorithm from \cite{van2016deep}.      

\subsection{Need for safety}\label{sec::NeedForSafety}
A key distinguishing component of RL when compared with other forms of machine learning is the trade-off between exploration and exploitation \cite{sutton1998reinforcement}. In order to achieve better reward an agent must try actions that it has not tried prior. Hence during the initial learning phase the agent will try all viable actions, unfortunately this curiosity may be fatal and can become expensive particularly when learning on a physical platform such as a robotic platform \cite{garcia2012safe}. For example while training a highway driving agent, use of unguided exploration could frequently lead to collision or near-miss scenarios, which may result in simulation resets, thus slowing the learning process. Additionally even after convergence, due to the function approximation by the $Q$-network, the trained agent may chose a non-safe maneuver. In order to address these issues, we augment the DDQN decision maker with an explicit \emph{short-horizon safety check} that is used both while learning and also during the implementation phase. Safe exploration or safe-RL is an active research topic, for a detailed survey on inclusion of various safety mechanisms during learning see  \cite{garcia2015comprehensive}.

\begin{figure}[!tbp]
  \begin{subfigure}[b]{0.5\textwidth}
    \includegraphics[width=\textwidth]{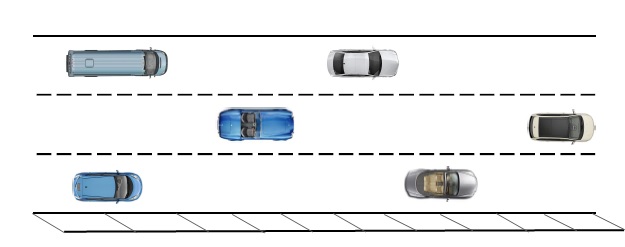}
    \caption{Highway with barricaded right lane.}
    \label{fig:BaricadedHighway}
  \end{subfigure}
  \hfill
  \begin{subfigure}[b]{0.5\textwidth}
    \includegraphics[width=\textwidth]{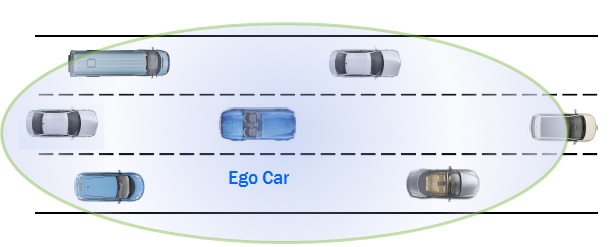}
    \caption{A three lane highway scenario.}
    \label{fig:EgoCar}
  \end{subfigure}
  \caption{An example for common sense road rule along with the ego vehicle perspective considered in this work.
  }
\end{figure}

A standard RL approach, such as DDQN, may require a lot of samples before learning that few actions could be potentially dangerous in certain states. For example, consider a highway where one of the lane has been barricaded by concrete blocks, see Fig.~\ref{fig:BaricadedHighway}. A standard epsilon-greedy algorithm may need to collide multiple times before learning that the action \emph{change lane to right} when in the right-most lane to be catastrophic. This may result in significant waste of learning effort as the agent will spend considerable time exploring irrelevant regions of the state and action space.  Additionally, by virtue of function approximation there is a small probability, where the trained DDQN agent may still chose \emph{change lane to right} leading to catastrophic outcome. This can be avoided by including an explicit short-horizon safety check that evaluates the action choice by the learning agent and provides an alternative safe action whenever it is feasible \cite{alshiekh2017safe}. 

In the RL algorithm presented in this work (see Section~\ref{sec::DQNArchitecture}) we use a simple safety check based on common sense road rules. This forces the agent to avoid non-safe actions in dangerous situations resulting in faster learning. Our approach is similar to a teacher who provides corrective action when it is necessary \cite{thomaz2008teachable}. Note that the safety controller may not be optimal, i.e., an expert in the task under consideration. Additionally, due to explicit safety check, new data can be obtained even in the implementation phase which can then be used for continuous adaption of the learned network, this is further elaborated in Section~\ref{sec::DQNArchitecture}.       

\section{Safe and adaptive decision making for AV} \label{sec::DQNArchitecture}

\begin{figure*}[htbp]
\centering
\includegraphics[width=\textwidth, height=4cm]{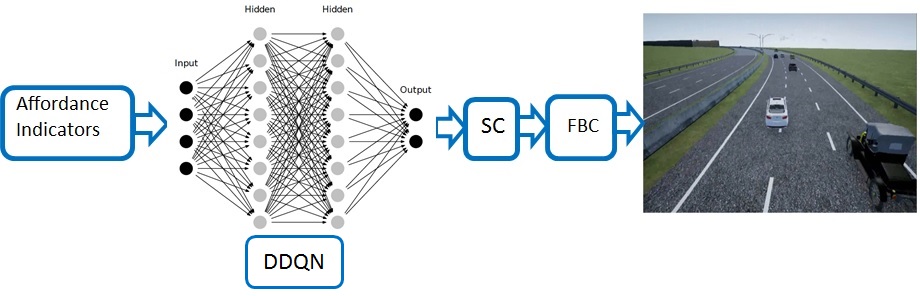}
\caption{DRL agent control architecture: \emph{SC} is the short-horizon safety check and \emph{FBC} is the low-level feedback controller.}
\label{fig:ControlDecisionMaking}
\end{figure*}

The DRL architecture used in this work for autonomous highway driving is given in Fig.~\ref{fig:ControlDecisionMaking}. 

Unlike the mediated perception method that relies on complete reconstruction of scene prior \cite{xu2017end,hecker2018end}, we use the concept of affordance indicators, i.e., state variables based on the direct perception approach from \cite{chen2015deepdriving}. For a three lane highway scenario, the ego vehicle (EV) can be surrounded by up to six traffic vehicles (TV), see Fig.~\ref{fig:EgoCar}. To define the traffic vehicle's variable we use the following notation
\begin{figure}[H]
\centering
\includegraphics[width=0.4\textwidth]{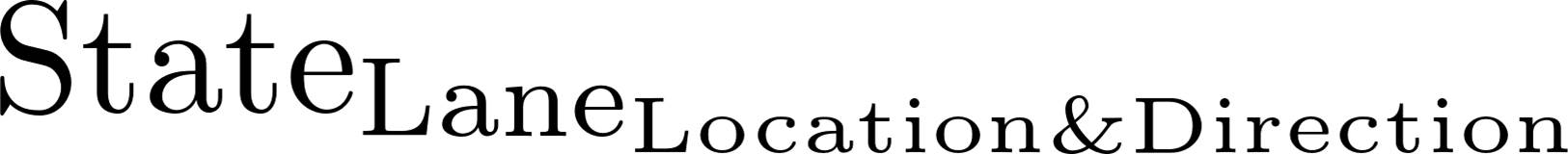}
\end{figure}
where State can be distance $d$ or velocity $v$, Lane is either right lane $rl$, center lane $cl$, or left lane $ll$. Location is either front $f$ or rear $r$ and Direction is either longitudinal $x$ or lateral $y$. 

In total the following $24$ indicators are used represent the spatio-temporal information of the six nearest traffic vehicles (see Fig.~\ref{fig:EgoCar}), they are formulated from the ego vehicle's perspective as:
 
\begin{itemize}
  \item The relative distance and velocity, in longitudinal and lateral direction, to the closest front car in center lane, i.e., $d_\mathrm{cl_{fx}}$, $v_\mathrm{cl_{fx}}$, $d_\mathrm{cl_{fy}}$, and $v_\mathrm{cl_{fy}}$. Similarly to the closest rear car in the center lane, i.e.,  $d_\mathrm{cl_{rx}}$, $v_\mathrm{cl_{rx}}$, $d_\mathrm{cl_{ry}}$, and $v_\mathrm{cl_{ry}}$. 
  \item  The relative distance and velocity, in longitudinal and lateral direction, to the closest front car in the right lane, i.e., $d_\mathrm{rl_{fx}}$, $v_\mathrm{rl_{fx}}$, $d_\mathrm{rl_{fy}}$, and $v_\mathrm{rl_{fy}}$. Similarly to the closest rear car in the right lane, i.e.,  $d_\mathrm{rl_{rx}}$, $v_\mathrm{rl_{rx}}$, $d_\mathrm{rl_{ry}}$, and $v_\mathrm{rl_{ry}}$. 
  \item  The relative distance and velocity, in longitudinal and lateral direction, to the closest front car in the left lane, i.e., $d_\mathrm{ll_{fx}}$, $v_\mathrm{ll_{fx}}$, $d_\mathrm{ll_{fy}}$, and $v_\mathrm{ll_{fy}}$. Similarly to the closest rear car in the left lane, i.e.,  $d_\mathrm{ll_{rx}}$, $v_\mathrm{ll_{rx}}$, $d_\mathrm{ll_{ry}}$, and $v_\mathrm{ll_{ry}}$.
\end{itemize}  
Here the lane occupancy of a TV and the closest car to EV are obtained using the methodology presented in \cite{zhang2017finite}. In addition to the 24 traffic vehicle states we use longitudinal velocity $v_\mathrm{e_x}$, lateral position $d_\mathrm{e_y}$, and lateral velocity $v_\mathrm{e_y}$, of the ego vehicle $e$. Since the affordance indicators are formulated w.r.t. ego vehicle we do not need a state corresponding to longitudinal position of the ego vehicle. These variables are minimal requirement for highway driving, however they are not sufficient for all highway driving tasks such as use of restricted lane, on-ramp to enter the highway, off-ramp to exit, etc.

A total of 27 affordance indicators, i.e., $s_t \in \mathbb{R}^{27}$ are used as an input to the deep $Q$-network. The $Q$-network is trained by using a customized variant of the deep double Q-learning algorithm from \cite{van2016deep}, see Algorithm~\ref{alg:DRL4AV}. For decision making of an AV in a highway driving scenario, we consider four action choices along longitudinal direction, namely, \emph{maintain}, \emph{accelerate}, \emph{brake}, and \emph{hard brake}. Whereas for lateral direction we assume three action choices, one for \emph{lane keep}, \emph{change lane to right}, and \emph{change lane to left}, respectively. This results in 12 unique action choices. For each of these action choices a numerical value can be assigned as in \cite{li2017game}, for example for longitudinal acceleration four different discrete choices  $\{a_1,0,-a_1, -a_2 \}$ may be considered. Alternatively, one can obtain the reference for the throttle or brake controller either using the intelligent driver model (IDM) \cite{treiber2000congested} or adaptive cruise control \cite{vahidi2003research}, etc. The reference for the steering controller is self-evident, it is either stay in lane or change lane to right or left. To obtain the steering command we use a simple feedback controller \cite{ivanovic2017lanechange}
\begin{equation}\label{eq:SteerCmd}
 \kappa_\mathrm{cmd} = f(k_\mathrm{road},e_\mathrm{y_{off}},e_\mathrm{\psi},T_\mathrm{LC},v_\mathrm{e_x})
\end{equation} 
where $k_\mathrm{road}$ is the road curvature, $e_\mathrm{\psi}$ is the heading angle offset, $T_\mathrm{LC}$ is the desired time to complete a  lane change, and the reference $e_\mathrm{y_{off}}$ is the lateral offset to the desired position. When the DDQN agent decides to perform a lane change, the absolute value of the lane offset $e_\mathrm{y_{off}}$ will be set to the lane width. The inputs to the steering feedback controller \eqref{eq:SteerCmd} can be considered as additional affordance indicators which can be obtained from the perception module. 
 
As elaborated in Section~\ref{sec::NeedForSafety}, we use an explicit short-horizon safety check to validate the action choice by DDQN. For the current action choice, the safety controller verifies common sense but well known rules  of the road such as ensuring a minimum relative gap to a TV based on its relative velocity,
\begin{equation}\label{eq:sfConstraint}
    d_\mathrm{TV}-T_\mathrm{min}*v_\mathrm{TV}>d_\mathrm{TV_{min}} 
\end{equation}    
where $d_\mathrm{TV}$, $v_\mathrm{TV}$ are the relative distance and relative velocity to a traffic vehicle, $T_\mathrm{min}$ is the minimum time to collision, $d_\mathrm{TV_{min}}$ is the minimum gap which must be ensured before executing the action choice by the  DDQN. If this condition is not satisfied and when it is feasible, an alternate safe action will be provided by the short-horizon safety controller. In this example we use a simple variant of the intelligent driver model (IDM) \cite{treiber2000congested} to provide safe alternative longitudinal action, and is formulated as
\begin{equation}\label{eq:sfCtrl} 
a_\mathrm{s} =
  \begin{cases}
    \text{Hard brake}       & \quad \text{if } T_\mathrm{C} \leq T_\mathrm{HB}\\
    \text{Brake}   & \quad \text{if } T_\mathrm{HB}<T_\mathrm{C} \leq T_\mathrm{B} \\
    \text{Maintain}   & \quad \text{if } T_\mathrm{B}<T_\mathrm{C}  
  \end{cases}
\end{equation}    
where $a_\mathrm{s}$ is the safe action, $T_\mathrm{C}$ is the calculated time to collision \cite{minderhoud2001extended}, $T_\mathrm{HB}$ and $T_\mathrm{B}$ are the thresholds above which the decision made by the DDQN agent is considered to be safe.

\begin{algorithm}[h]
	\begin{algorithmic}[1]
		\STATE Initialize: $\mathrm{Buf}_\mathrm{S}$,  $\mathrm{Buf}_\mathrm{C}$, $Q(\theta)$  and target network $\hat{Q}(\hat{\theta})$ with $\hat{\theta}=\theta$ \\		
        \FOR{episode =1,$N_\mathrm{e}$,}
		\STATE Initialize: $\{1,\cdots ,N_\mathrm{T}\}$ cars randomly, obtain affordance indicator $s_0$\\
		\FOR{samples $t$ =1,$N_\mathrm{S}$, or Collision, }
    		\STATE With $\epsilon$ select random action $a_t$, else $a_t = \arg\max_\mathrm{a} Q\left((s_t,a,\theta_t) \right)$\\
            \STATE For ego car: \textbf{If} $a_t$ is not safe \textbf{Then} store $\left(s_t,a_t,*,r_\mathrm{col}\right)$ in $\mathrm{Buf}_\mathrm{C}$ and replace $a_t$ by safe action $a_\mathrm{s}$ \\
    		\STATE Apply action, observe $s_{t+1}$ and obtain $r_{t+1} = \rho(s_t, s_{t+1}, a_{t})$ \\
            \IF{Collision}
                \STATE Store transition $\left(s_t,a_t,*,r_\mathrm{col}\right)$ in collision buffer $\mathrm{Buf}_\mathrm{C}$ \\
            \ELSE
                \STATE Store transition $\left(s_t,a_t,s_{t+1},r_{t+1}\right)$ in safe buffer $\mathrm{Buf}_\mathrm{S}$ \\
            \ENDIF
            \STATE Sample random minibatch $\left(s_j,a_j,s_{j+1},r_{j+1}\right)$ from $\mathrm{Buf}_\mathrm{S}$ and $\mathrm{Buf}_\mathrm{C}$
            \STATE Set \begin{equation*}
                         y_j =
                                  \begin{cases}
                                    r_{j+1}       \quad \text{if sample is from } \mathrm{Buf}_\mathrm{C}\\
                                    r_{j+1}+\gamma \hat{Q} \Big( s_{j+1},  \arg\max_a Q\left(s_{j+1},a,\theta_t\right) \\
\hspace{2cm} ,\hat{\theta}_t \Big) \quad \text{if sample is from } \mathrm{Buf}_\mathrm{S}
                                  \end{cases}
                        \end{equation*}
            \STATE Perform gradient descent on $\left(y_j-Q(s_j,a_j,\theta_t)\right)^2$ w.r.t. $\theta$
            \STATE Every $N_\mathrm{C}$ episodes set $\hat{Q} = Q$
    		\ENDFOR
		\ENDFOR
	\end{algorithmic}
	\caption{A DRL based safe decision maker for autonomous highway driving}
	\label{alg:DRL4AV}
\end{algorithm}

In particular we use the following safety check prior to performing an action given by DDQN: 
\begin{enumerate}
\item Instead of the in-lane longitudinal action by DDQN, chose a safe action using \eqref{eq:sfCtrl} if \eqref{eq:sfConstraint} is not satisfied and the ego vehicle is faster than the preceding vehicle.
\item If the ego vehicle is in left most lane then \emph{change lane to left} is not valid, similarly for the right lane. 
\item For \emph{change lane to left} continuously monitor \eqref{eq:sfConstraint} for preceding car, the front and the rear car in the target lane. If condition \eqref{eq:sfConstraint} fails then  lane change is either not initiated or aborted, similarly for the \emph{change lane to right}. Lane abort is initiated by switching to an appropriate action which is opposite to the current lateral direction. 
\end{enumerate}

Generally in many RL applications after learning, the trained agent is frozen and used as a feedback controller. However, in reality the agent may encounter new information, additionally there can be a considerable variation between the training environment and the real-world experience. Also due of function approximation there can be a small probability of choosing an unsafe action even by the trained agent, this can happen even after convergence and in the absence of any explicit exploration. In order to address these issues, in the implementation phase we augment the trained DDQN agent with the short-horizon safety check that was used during learning. Any new safety violation data will be added to the collision buffer $\mathrm{Buf}_\mathrm{C}$,  by using the training part of the Algorithm~\ref{alg:DRL4AV} (line 13 until 15) the learned agent can be re-trained or adapted in a continuous manner. In the following section we apply the developed DRL based decision making Algorithm~\ref{alg:DRL4AV} for autonomous highway driving.

\section{Use of DRL for decision making}\label{sec::evalLearnedControl}

In this section we will show the usability of our DRL based decision making Algorithm~\ref{alg:DRL4AV} for autonomous highway driving. First, we introduce the vehicle dynamical model that was used training, following this we will elaborate on the training environment and evaluate the learned policy. 

\subsection{Vehicle dynamics}
Each vehicle is modeled using a computationally efficient point-mass model. For longitudinal equations of motion we use a discrete-time double integrator,
\begin{align}\label{eq1:LogiModel}
x(t+1) &= x(t)+v_{x}(t) \Delta t \nonumber \\ 
v_{x}(t+1) &= v_{x}(t) + a_\mathrm{x}(t) \Delta t.
\end{align}
where $t$ is the time index, $\Delta t$ is the sampling time, $x\in \mathbb{R}$ is the longitudinal position, and $v_{x} \in \mathbb{R}$ is the longitudinal velocity of the vehicle. For the lateral motion we assume a simple kinematic model 
\begin{equation} \label{eq2:LatModel}
y(t+1) = y(t)+v_\mathrm{y}(t) \Delta t
\end{equation}
where $y \in \mathbb{R}$ is the lateral position of the car. In \eqref{eq1:LogiModel} and \eqref{eq2:LatModel}, the external control inputs $a_\mathrm{x}(t)$ and $v_\mathrm{y}(t)$ represents the longitudinal acceleration and lateral velocity of the vehicle, respectively.

We assume $a_\mathrm{x}(t)$ to vary from nominal acceleration,  to hard brake and is discretized into four values, i.e., $a_\mathrm{x} = \{a_1,0,-a_1, -a_2 \}$, with $a_1 = 2\, m/s^2$ and $a_2 = 4\, m/s^2$. Only in case of emergency hard braking of $a_\mathrm{x}=-a_2$ is applied. The lateral velocity $v_\mathrm{y}(t)$ provides a reference lane for the vehicle, we assume a lane change action requires 5 seconds to complete \cite{toledo2007modeling}, with an option of aborting the lane change maneuver during each sampling instance. In this work we use $1 \,Hz$ sampling. 

\subsection{Simulation environment}

\begin{figure}[htbp]
\centering
\includegraphics[width=0.48\textwidth]{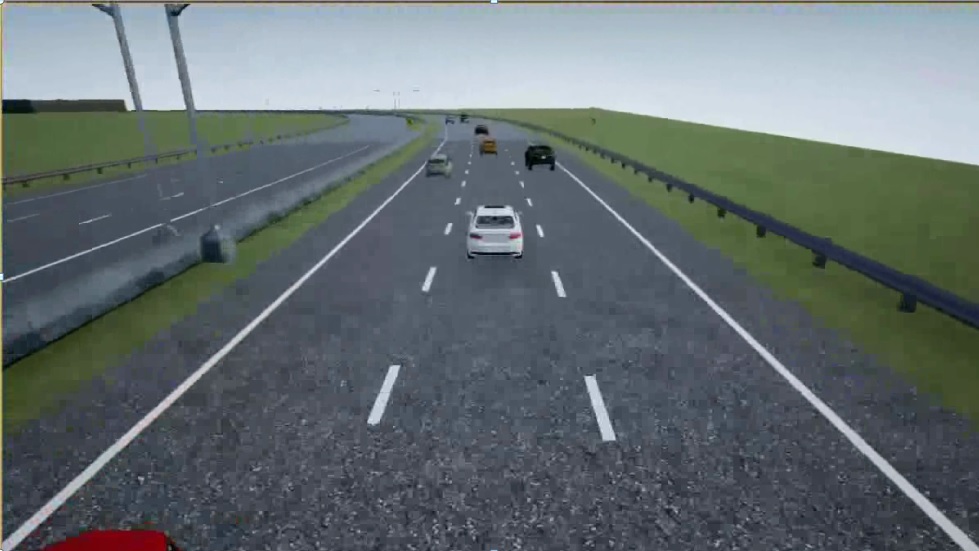}
\caption{A schematic of the simulation environment used for training.}
\label{fig:SimulationEnv}
\end{figure}

A schematic of the simulation environment used for training is given in Fig.~\ref{fig:SimulationEnv}. It is a three lane circular loop and is used to approximate an infinite stretch of straight highway. At the beginning of an episode, anywhere between $\{1,\cdots,N_\mathrm{T}\}$ number of cars are placed randomly within a distance of $250\, m$ from the ego car. In this example we chose $N_\mathrm{T}$ to be 30. 
 
During learning stage the ego car (for e.g., white Fusion in Fig.~\ref{fig:SimulationEnv}) uses an $\epsilon$-greedy policy to make decisions, whereas for the traffic vehicles a combination of controllers from \cite{li2017game} and \cite{zhang2017finite} are used along with an IDM controller. Additionally the traffic vehicles can randomly chose to perform lane change. For the traffic vehicles, the system parameters such as maximum velocity are randomly chosen. This is to ensure a diverse traffic scenario in training and evaluation. We assume that all the traffic vehicles take into account the relative distance and velocity to preceding vehicle before making a decision, i.e., they will not rear end the preceding car in the same lane. We use Algorithm~\ref{alg:DRL4AV} to train an agent for decision making.
   
\subsection{Decision making for ego vehicle}
In order to train the policy $\pi$ we use a reward function $\rho$ that consists of a set driving goals for the ego car. It is formulated as a function of 
\begin{itemize}
\item Desired traveling speed subject to traffic condition \eqref{eq:rewv},
\item Desired lane and lane offset subject to traffic condition \eqref{eq:rewy},
\item Relative distance to the preceding car based on relative velocity \eqref{eq:rewx},
\end{itemize}

\begin{align}
r_v &= e^{-\frac{\left(v_\mathrm{e_x}-v_\mathrm{des}\right)^2}{10}}  -1, \label{eq:rewv} \\
r_y &= e^{-\frac{\left(d_\mathrm{e_y}-y_\mathrm{des}\right)^2}{10}}-1, \label{eq:rewy} \\
r_x &= \begin{cases}
    e^{-\frac{\left(d_\mathrm{lead}-d_\mathrm{safe}\right)^2}{10 d_\mathrm{safe}}}-1       & \quad \text{if } e_x < d_\mathrm{safe}\\
    0  & \quad \text{otherwise,}
  \end{cases} \label{eq:rewx}
\end{align}
where $v_\mathrm{e_x}$, $d_\mathrm{e_y}$, and $d_\mathrm{lead}$ are the ego velocity, lateral position, and the longitudinal distance to the lead vehicle respectively. Similarly, $v_\mathrm{des}$, $y_\mathrm{des}$, and $d_\mathrm{safe}$ are the desired speed, lane position, and safe longitudinal distance to the lead vehicle respectively.

\begin{figure}[h!]
\centering	
  \begin{subfigure}[htbp]{0.48\textwidth}
    \includegraphics[width=\textwidth]{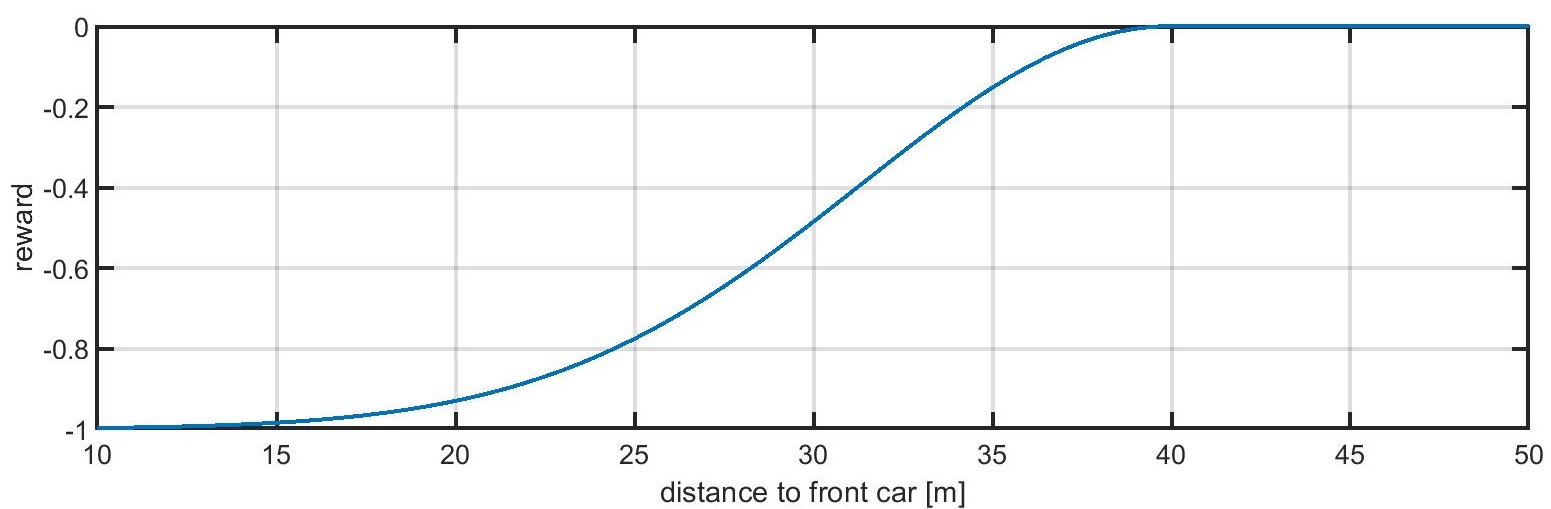}
    \caption{Reward for desired relative distance.}
    \label{fig:rewD}
  \end{subfigure}  
  \quad  
  \begin{subfigure}[htbp]{0.48\textwidth}
    \includegraphics[width=\textwidth]{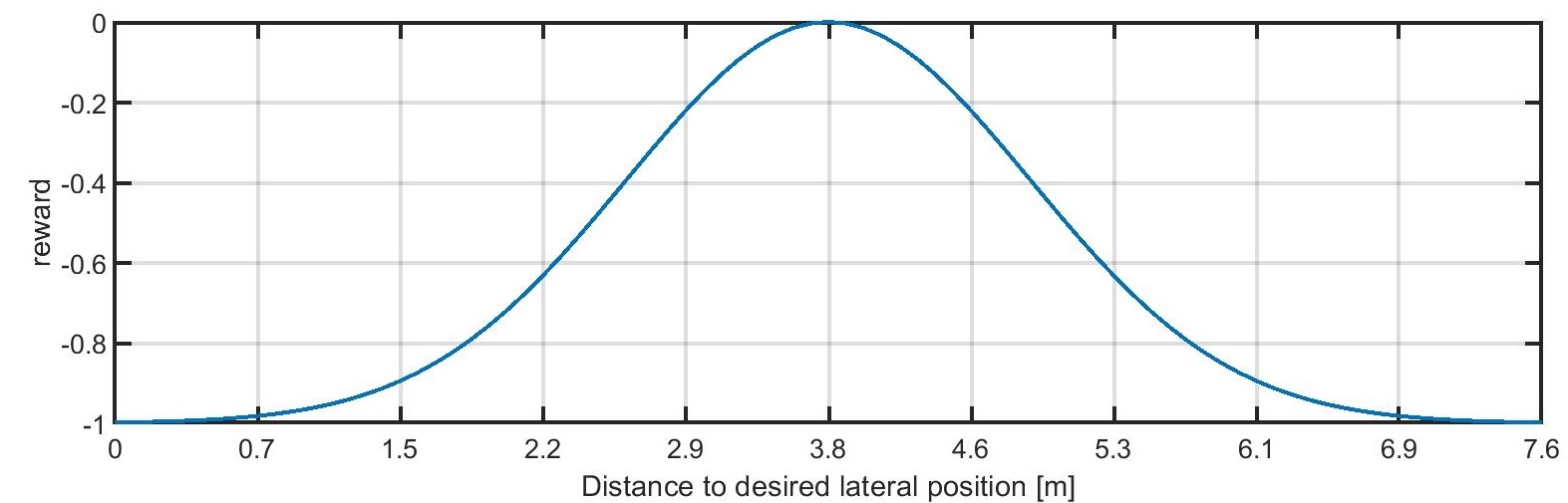}
    \caption{Reward for desired lateral position.}
    \label{fig:rewY}
  \end{subfigure}
  
  \begin{subfigure}[htbp]{0.48\textwidth}
    \includegraphics[width=\textwidth,height=3cm]{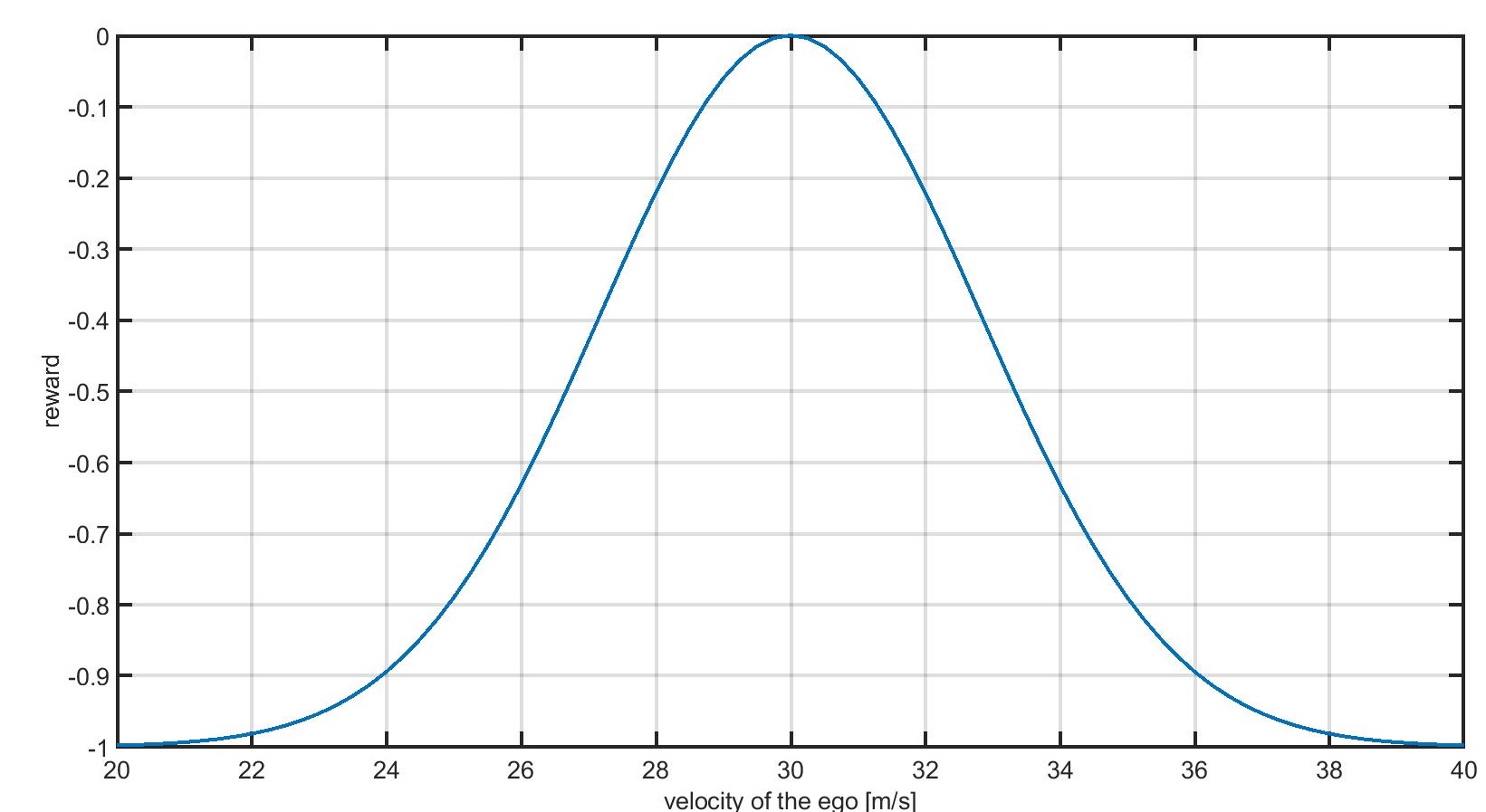}
    \caption{Reward for desired ego speed.}
    \label{fig:rewV}
  \end{subfigure}
  
  \caption{Reward for the ego car based on traffic condition, sub goals are weighted equally when calculating the final reward.}
  \label{fig:rewardFunc}
\end{figure}

Fig.~\ref{fig:rewardFunc} gives an indicative plot of the reward functions \eqref{eq:rewv}-\eqref{eq:rewx}, it is formulated assuming $v_\mathrm{des}\,=\,30\; m/s$ which can be achieved in the center lane i.e., $y_\mathrm{des}\,=\,3.8 m$ with a minimum safe distance $d_\mathrm{safe}\,=\,40m$. The desired values are based on the traffic condition and can change depending on the scenario. For slow/fast moving traffic the peak in Fig.~\ref{fig:rewV} will be adjusted based on the traffic condition. In this work we penalize the ego vehicle if it cannot maintain a minimum time headway of at least $1.3$ seconds.

\begin{figure}[htbp]
\centering
\includegraphics[width=0.48\textwidth]{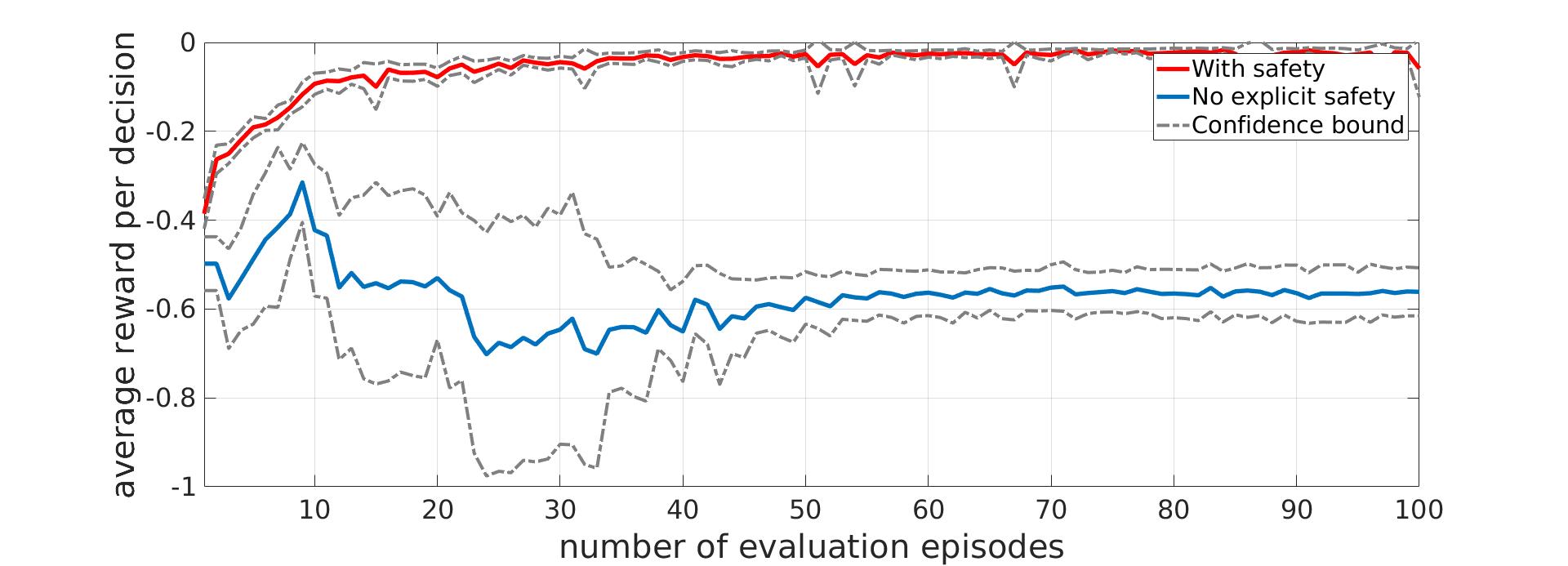}
\caption{Average learning curve with confidence bound for with and without short horizon safety check in Algorithm~\ref{alg:DRL4AV}.}
\label{fig:AvRewPerEp}
\end{figure}

During learning, we evaluate the (partially) trained DRL controller  every $100^{\text{th}}$ episode. Fig.~\ref{fig:AvRewPerEp} shows the average reward per decision during the training phase. It takes nearly $2000$ episodes for the agent to converge. We train the DRL agent for a total of $10000$ episodes. Where each episode lasts until $200$ samples or collision, whichever is earlier. Exploration is continuously annealed from $1$ to $0.2$  over first $7000$ episodes and then kept constant for the remaining duration of learning. The $Q$-network is a deep neural network with $2$ hidden layers each having $100$ fully connected leaky ReLU's \cite{maas2013rectifier}. We train the network using Adam optimizer \cite{kingma2014adam} with a fixed learning rate of $1e-4$. 

For the highway driving task, the safety controller was found to be a key component for learning a meaningful policy. Fig.~\ref{fig:AvRewPerEp} shows the mean and confidence bound for training with and without safety controllers over $200$ training iterations of Algorithm~\ref{alg:DRL4AV}. Training a standard DDQN agent without explicit safety check could not learn a decent policy and always resulted in collision. Whereas DDQN with explicit safety check was able to converge to an optimal policy. Based on \eqref{eq:rewv} - \eqref{eq:rewx}, the maximum reward an agent can receive is zero per decision, the average reward per decision obtained by our trained DDQN agent with safety check is around $-0.025$.  
  
\begin{figure}[htbp]
\centering
\includegraphics[width=0.48\textwidth]{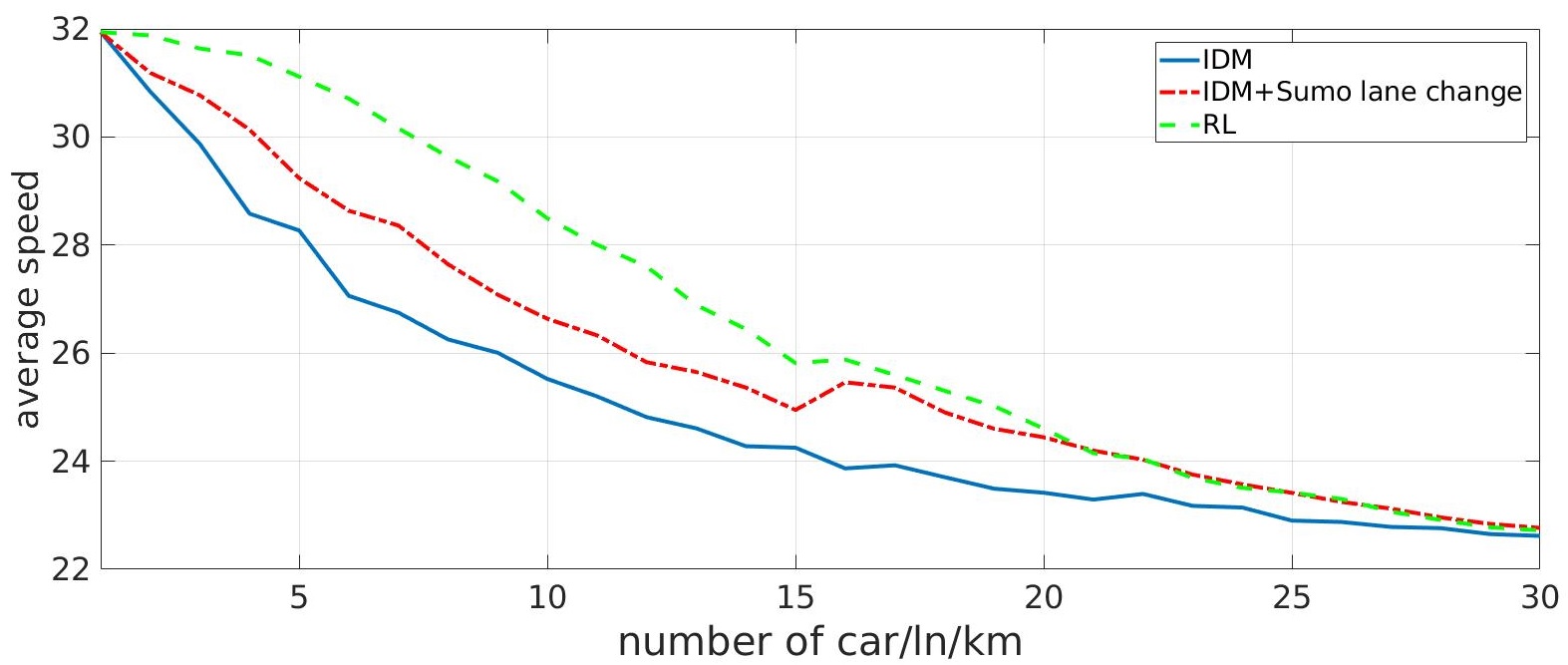}
\caption{Average speed for simple IDM controller, with lane change, and trained RL agent.}
\label{fig:Avspd}
\end{figure}

In Fig.~\ref{fig:Avspd} We evaluate our trained DDQN agent to obtain average velocity with increase in traffic density.  We compare this against modified safety controller from  \eqref{eq:sfCtrl}, the modification provides an acceleration command when the calculated time to collision $T_\mathrm{C}$ is higher than $T_\mathrm{A}$. This is referred as IDM in Fig.~\ref{fig:Avspd}. It must be noted IDM controller from \eqref{eq:sfCtrl} cannot initiate lane change, in order to address this
we integrate IDM with lane change decision making from \cite{kesting2007general} and \cite{erdmann2014lane}. Fig.~\ref{fig:Avspd}, clearly demonstrates advantage of RL for high level decision making when compared to model-based approaches. With the increase in traffic density both the trained DDQN agent and the model-based lane change controller converges to IDM controller. This is anticipated since lane change is neither safe nor advantageous in higher traffic density.  
  
Use of two explicit buffers namely $\mathrm{Buf}_\mathrm{S}$ and $\mathrm{Buf}_\mathrm{C}$ in Algorithm~\ref{alg:DRL4AV} to store safe and non-safe transitions is simplified version of prioritized experience reply (PER) from \cite{schaul2015prioritized}. Fig.~\ref{fig:CompAdap} shows the mean and confidence bound for training with two buffers and PER over $200$ training iterations of Algorithm~\ref{alg:DRL4AV}. For the highway driving example using two explicit buffers provides marginally better policy when compared to PER.  This can be due to clear bifurcation of safe and non-safe transitions. 
\begin{figure}[htbp]
\centering
\includegraphics[width=0.48\textwidth]{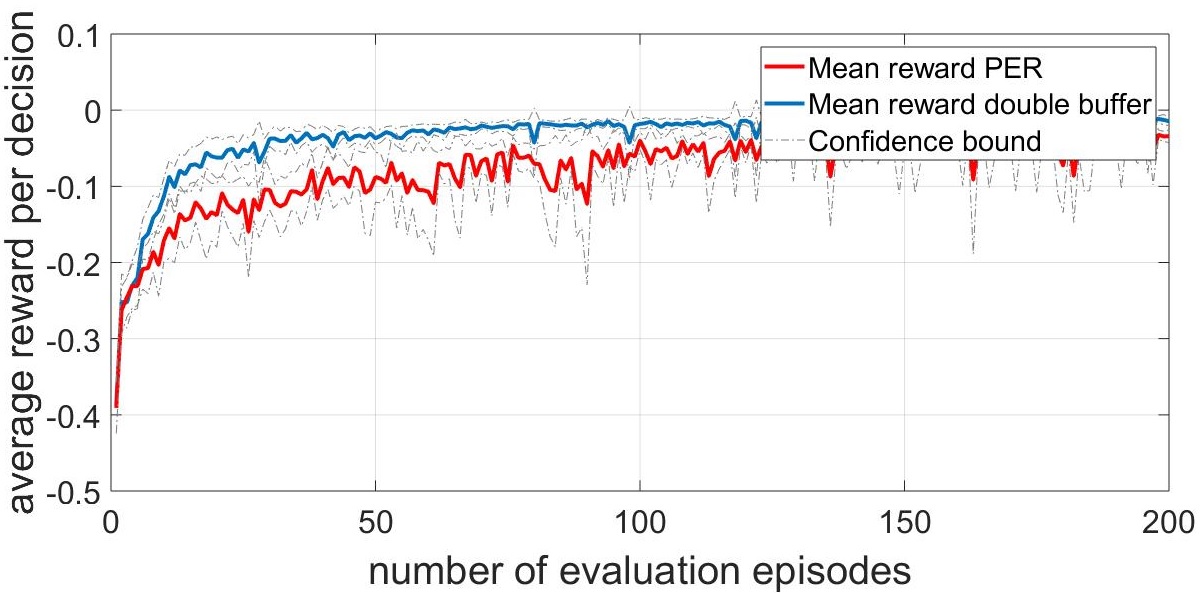}
\caption{Mean learning curve with confidence bound for Algorithm~\ref{alg:DRL4AV} and prioritized experience reply \cite{schaul2015prioritized}. In this work we used the PER implementation from \cite{stable-baselines}.}
\label{fig:CompAdap}
\end{figure}

\subsection{Continuous adaptation}
During the implementation phase, we replace the $\epsilon$-greedy policy, i.e., $\pi_\epsilon$ in Algorithm~\ref{alg:DRL4AV} line 5 by the learned policy $\pi$. Whenever the control decision by DDQN fails the short-horizon safety check, buffer $\mathrm{Buf}_\mathrm{C}$ is updated with additional data. Using a lower learning rate than the one used for training, $Q$-network can be retrained (line 13 until 16). Fig.~\ref{fig:CompAdap} shows the continuous adaptation result over 30K episodes and is obtained by averaging the data over 10k episodes using a moving average filter. Because of filtering, the mean number of safety trigger increases over first 10k episodes and stays constant for no adaptation scenario whereas it monotonically decreases to a smaller value thanks to continuous adaptation. Even with continuous adaptation the mean number safety trigger never converges to zero, this may be due to 
\begin{enumerate}
    \item Use of function approximation where a trained NN can potentially chose a non-safe action,
    \item Use of rigid and static safety rules. 
\end{enumerate}
In our future work we plan to address this issue. 

\begin{figure}[htbp]
\centering
\includegraphics[width=0.48\textwidth]{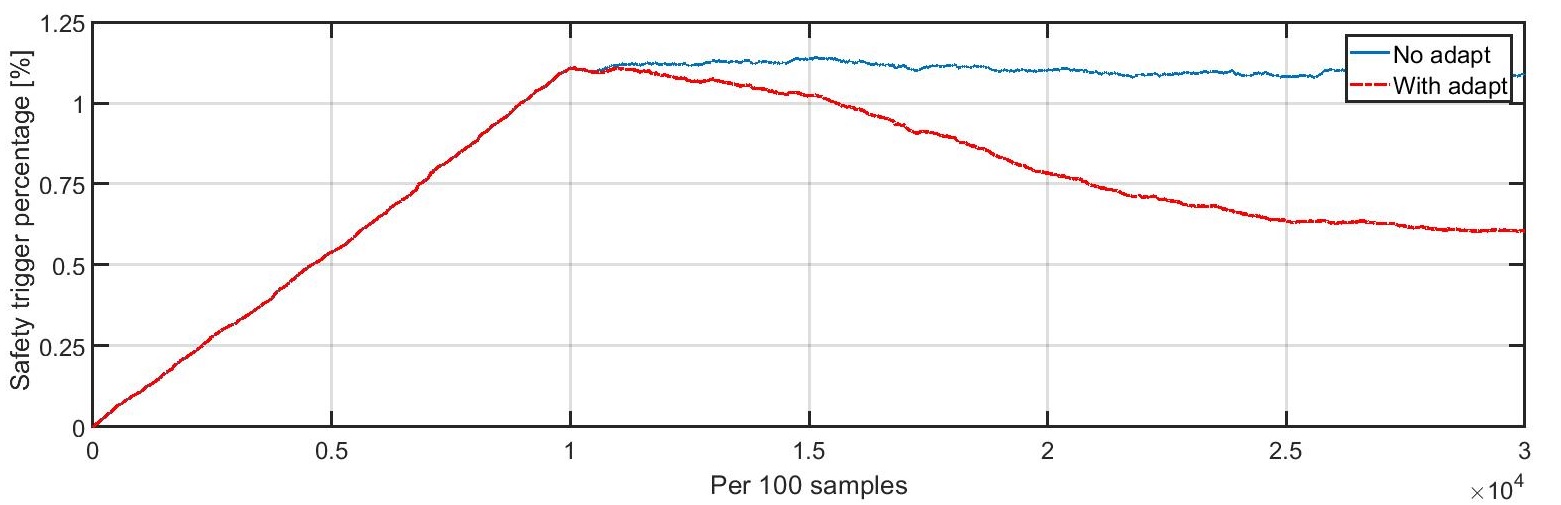}
\caption{Comparison of number of safety trigger after learning with and without continuous adaptation.}
\label{fig:CompAdap}
\end{figure}
 
\section{Conclusions}\label{sec:conclusion}
In this paper, we presented a control architecture based deep RL framework for safe decision making in autonomous driving. The $Q$-network is augmented with a short-horizon safety control thus embedding few common rules of the road into AV decision making. By using well known feedback controller, high level decision by DDQN is converted into a low-level actuation vis-\`{a}-vis, throttle, brake, and steering. We have evaluated the learned controller under varying traffic density, the results demonstrate the superior capabilities of the learned DRL agent. Finally, by modifying the learning algorithm we have demonstrated the continuous adaptation framework that has shown to reduce the number of safety triggers. In our future work we plan to extend this work to achieve mandatory lane change, and to perform on-ramp to off-ramp highway driving.

\section*{Acknowledgment}
The authors would like to thank Dr. Vladimir Ivanovic for providing the low-level steering controller.

\section*{Appendix}
Table~\ref{tab::parameterVal} lists the parameter values used in this work.

\begin{table}
\begin{center}

 \begin{tabular}{ | l | l | l |}
 \hline
 Symbol & Description & Value \\ \hline
 $\gamma$ & Discount factor & 0.9 \\ \hline
 $\Delta t$ & Sampling time & 1 sec \\ \hline
 $T_\mathrm{LC}$ & Lane change duration & 5 sec \\ \hline
 $T_\mathrm{min}$ & Minimum time to collision & 3 sec \\ \hline
 $T_\mathrm{HB}$ & Hard break threshold & 2 sec \\ \hline
 $T_\mathrm{B}$ & Nominal break threshold & 3 sec \\ \hline
 $T_\mathrm{A}$ & Nominal acceleration threshold & 8 sec \\ \hline
 $d_\mathrm{TV_\mathrm{min}}$ & Minimum gap & 15 m \\ \hline
 \end{tabular}
 \end{center}
\caption{Parameter values used in this work.}
\label{tab::parameterVal}
\end{table}
 
\bibliographystyle{IEEEtran}
\bibliography{root}

\end{document}